\documentclass[letterpaper, 10 pt, conference]{ieeeconf}  

\usepackage[utf8]{inputenc}

\IEEEoverridecommandlockouts                              

\usepackage[american]{babel}

\usepackage{booktabs} 
\usepackage{amssymb}
\usepackage{amsmath}
\usepackage{tikzit}
\usepackage{url}

\tikzstyle{black dot}=[fill=black, draw=black, shape=circle]
\tikzstyle{red dot}=[fill=red, draw=black, shape=circle]
\tikzstyle{green dot}=[fill={rgb,255: red,44; green,141; blue,29}, draw=black, shape=circle]
\tikzstyle{red unfilled}=[fill=white, draw=red, shape=circle, inner sep=2 pt]
\tikzstyle{yellow unfilled}=[fill=white, draw={rgb,255: red,214; green,214; blue,0}, shape=circle, inner sep=2 pt]
\tikzstyle{yellow dot}=[fill={rgb,255: red,247; green,247; blue,0}, draw=black, shape=circle]
\tikzstyle{small black dot}=[fill=black, draw=black, shape=circle, inner sep=2pt]
\tikzstyle{black rectangle}=[fill=white, draw=black, shape=rectangle, minimum width=1, minimum height=1]
\tikzstyle{small grey circle}=[fill={rgb,255: red,191; green,191; blue,191}, draw={rgb,255: red,191; green,191; blue,191}, shape=circle, inner sep=2pt]
\tikzstyle{small green dot}=[fill={rgb,255: red,44; green,141; blue,29}, draw=black, shape=circle, inner sep=2pt]
\tikzstyle{small red dot}=[fill=red, draw=black, shape=circle, inner sep=2 pt]
\tikzstyle{small yellow dot}=[fill={rgb,255: red,247; green,247; blue,0}, draw=black, shape=circle, inner sep=2 pt]
\tikzstyle{smol black dot}=[fill=white, draw=black, shape=circle, inner sep=1 pt]
\tikzstyle{tiny black dot}=[fill=black, draw=black, shape=circle, inner sep=1pt]
\tikzstyle{new style 0}=[fill=white, draw=black, shape=circle]
\tikzstyle{tiny grey dot}=[fill={rgb,255: red,191; green,191; blue,191}, draw={rgb,255: red,191; green,191; blue,191}, shape=circle, inner sep=1 pt]

\tikzstyle{green edge}=[-, draw={rgb,255: red,44; green,141; blue,29}]
\tikzstyle{red edge}=[draw=red, -]
\tikzstyle{green edge thick}=[-, draw={rgb,255: red,44; green,141; blue,29}, line width=6pt]
\tikzstyle{new edge style 0}=[-, dashed]
\tikzstyle{red arrow}=[->, draw=red]
\tikzstyle{green arrow}=[->, draw={rgb,255: red,44; green,141; blue,29}]
\tikzstyle{black arrow}=[->, thick]
\tikzstyle{dotted red}=[-, dashed, draw=red, thick]
\tikzstyle{blue edge}=[-, draw={rgb,255: red,106; green,126; blue,255}]
\tikzstyle{orange edge}=[-, draw={rgb,255: red,255; green,175; blue,76}]
\tikzstyle{thick blue}=[-, draw=blue, thick]
\tikzstyle{thick green}=[-, draw={rgb,255: red,0; green,159; blue,0}, thick]
\tikzstyle{thick red}=[-, thick, draw=red]
\tikzstyle{thick gray}=[-, draw={rgb,255: red,128; green,128; blue,128}, thick]
\tikzstyle{gray arrow}=[->, draw={rgb,255: red,211; green,211; blue,211}]
\tikzstyle{thick dotted green}=[-, draw={rgb,255: red,44; green,141; blue,29}, dashed, thick]
\tikzstyle{new edge style 1}=[-]
\tikzstyle{thin black arrow}=[->]
\tikzstyle{thin gray line}=[-, draw={rgb,255: red,214; green,214; blue,214}]
\tikzstyle{new edge style 2}=[draw={rgb,255: red,0; green,0; blue,212}, ->]

\usepackage{balance}

\title{\LARGE \bf
FeatSense -- A Feature-based Registration Algorithm with GPU-accelerated TSDF-Mapping Backend for NVIDIA Jetson Boards
}

\author{Julian Gaal$^{1}$, Thomas Wiemann$^{2,3}$, Alexander Mock$^{1}$, and Mario Porrmann$^{1}$ 
\thanks{$^{1}$Institute of Computer Science, Osnabrück University, 49069 Osnabrück, Germany
        {\tt\small firstname.lastname@uos.de}}%
\thanks{$^{2}$Fulda University of Applied Sciences, Applied Computer Science, 36037 Fulda, Germany
        {\tt\small thomas.wiemann@informatik.fh-fulda.de}}%
\thanks{$^{3}$German Research Center for Artificial Intelligence (DFKI), Plan-based Robot Control, 49080 Osnabrück, Germany
        {\tt\small thomas.wiemann@dfki.de}}%
}

\begin{document}

\maketitle
\thispagestyle{empty}
\pagestyle{empty}

\begin{abstract}

This paper presents FeatSense, a feature-based GPU-accelerated SLAM system for high resolution LiDARs, combined with a map generation algorithm for real-time gener\-ation of large Truncated Signed Distance Fields (TSDFs) on embedded hardware.
FeatSense uses LiDAR point cloud features for odometry estimation and point cloud registration.
The registered point clouds are integrated into a global Truncated Signed Distance Field (TSDF) representation. 
FeatSense is intended to run on embedded systems with integrated GPU-accelerator like NVIDIA Jetson boards.
In this paper, we present a real-time capable TSDF-SLAM system specially tailored for close coupled CPU/GPU systems. 
The implementation is evaluated in various structured and unstructured environments and benchmarked against existing reference datasets. 
The main contribution of this paper is the ability to register up to 128 scan lines of an Ouster OS1-128 LiDAR at 10Hz on a NVIDIA AGX Xavier while achieving a TSDF map generation speedup by a factor of \emph{100} compared to previous work on the same power budget.

\end{abstract}

\section{INTRODUCTION}

In previous work, we presented HATSDF-SLAM~\cite{eisoldt2022fully} for LiDAR-based SLAM that maintains a global truncated signed distance field (TSDF) of the environment while performing localization directly within this representation. 
It is implemented on a low-power embedded system with an integrated Field Programmable Gate Array (FPGA) to achieve localization in real time on slowly moving systems.
This FPGA-based Point-to-TSDF registration allows real-time SLAM for LiDARs with up to 16 horizontal scan lines at moderate walking speeds and 32 horizontal scan lines at very slow speeds.
HATSDF offers good registration performance, especially in indoor environments, but the implementation is effectively limited by two interconnected factors: registration run time and local map size.
The TSDF representation is stored in a local map that is incrementally shifted after a movement threshold. 
During the shift operation, newly mapped parts are written to disk while previously mapped areas are loaded into the local map in a very disk I/O heavy operation. 

Map shift is limited by disk I/O speed, while the registration is limited by DRAM accesses to a local segment of the map. 
Increasing the FPGA's power consumption may lead to faster computation but makes the system less suitable for long term mapping on devices with limited battery capacity.
Also, due to the limited number of available memory ports, scalability to higher resolution is problematic, as the algorithm is highly memory bound.
On the other side, embedded systems with GPUs like the NVIDIA Jetson boards are becoming largely available and offer GPU-based acceleration on the device while still keeping the power consumption low.
The main advantage of such devices is that they offer more memory and higher bandwidths compared to a FPGA system.
In this paper we present FeatSense\footnote{\url{https://github.com/juliangaal/warpsense}}, a GPU-accelerated, high-resolution mobile mapping system for start-of-the-art LiDARs that overcomes the bandwidth bottlenecks and speed limitations of HATSDF SLAM.
We show that our solution works well in different environments and allows higher moving speed than the previous system.
A comparison of our approach with an existing LOAM algorithm shows that our implementation is able to generate competitive results in terms of accuracy, while improving the supported scanning resolution and drastically reducing computation time. 

\section{RELATED WORK}

\begin{figure*}
  \centering
  \includegraphics[width=0.49\linewidth]{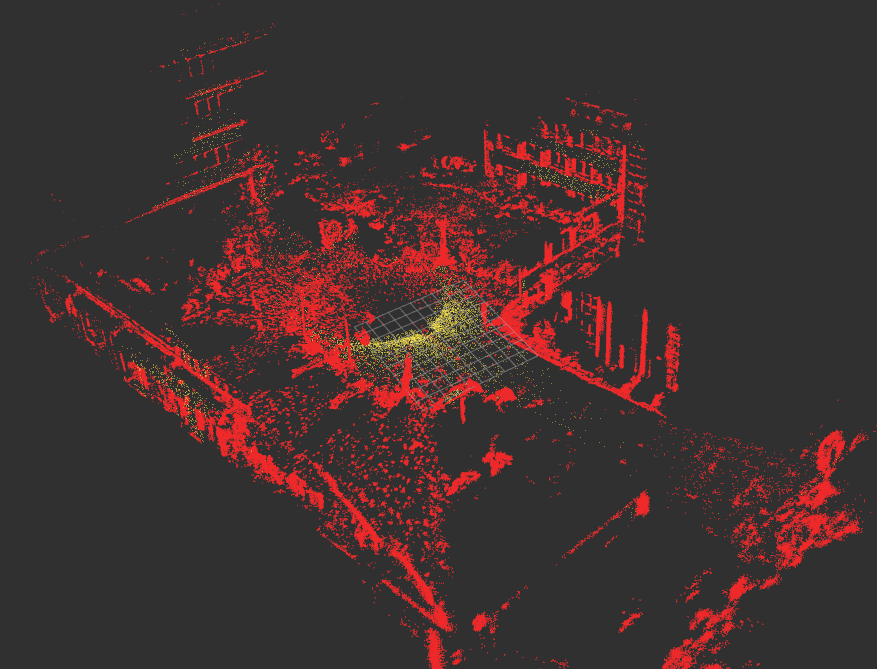} 
  \hspace{1mm}
  \includegraphics[width=0.49\linewidth]{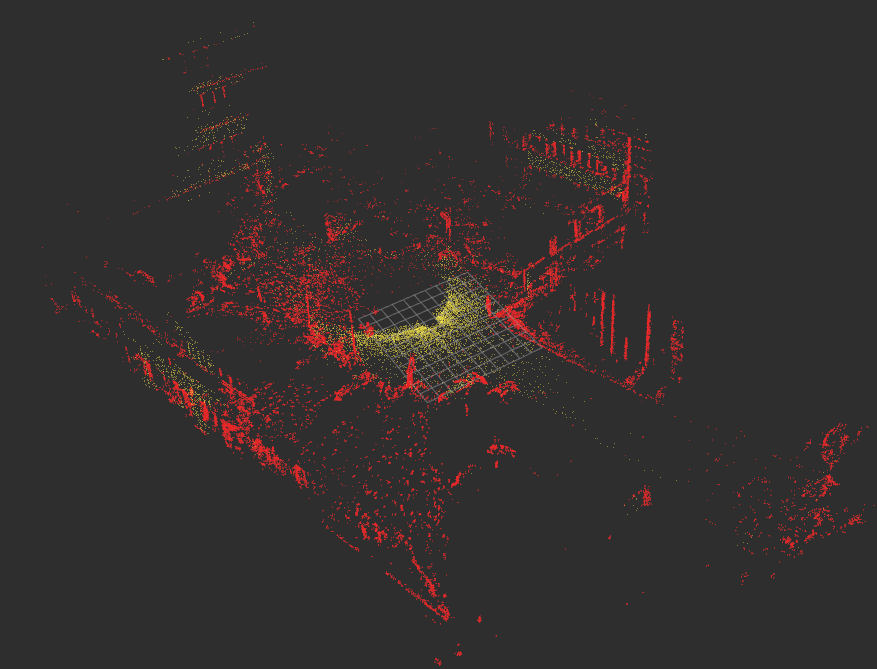}
  \caption{Edge map without integration of the depth image (left) and after filtering out edges that to not exist in both domains (right) results in less but more stable feature points.} 
  \label{fig:filtering}
\end{figure*}

LiDAR Odometry And Mapping (LOAM)~\cite{LOAM} uses deceptively simple feature extraction, smart correspondence matching and outlier rejection to estimate the ego-motion of a LiDAR to build 3D point clouds.
Feature points are computed on horizontal scan lines, which requires a structured point cloud. 
A structured point cloud, sometimes referred to as ordered, allows access by row $P$ and column $R$ in a $P\times R$ scan, where each horizontal scan line refers to scan line $P_k$ at a specific angle $\alpha$ from the sensor origin. 
If a LiDAR does not provide a structured point cloud, it can be sorted by the vertical angle to the emitter to reproduce an ordered structure.
The invention of the original LOAM algorithm ushered in a wave of feature-based SLAM algorithms for such LiDARs. 
A-LOAM~\cite{ALOAM} laid the basis from many LOAM-based SLAM solutions. 
It uses of the Ceres optimization library~\cite{CERES} to significantly speed up the scan matching optimization. 
Ceres allows automatic and numeric differentiation, is highly optimized and offers modern multithreaded linear algebra routines, including multiple solvers for several loss functions. 
The library is explicitly designed to make it easy for users to modify the objective function without worrying about the optimization process itself.
LEGO-LOAM~\cite{LEGOLOAM} is a variation of LOAM, that performs ground segmentation and feature clustering for more accurate correspondence estimation in ground-based robot applications. 
It separates the scan matching task into two distinct problems that are solved separately in a two-step optimization scheme for position $t$ and orientation $\theta$.
The system's pose [$t_z$, $\theta_{roll}$, $\theta_{pitch}$] is obtained by matching the surface features extracted from ground segmentation, followed by the estimation of [$t_z$, $t_y$, $\theta_{yaw}$] with detected edge features while maintaining [$t_z$, $\theta_{roll}$, $\theta_{pitch}$] as a constraint.
LIO-SAM~\cite{LIOSAM} optimizes both odometry estimation by LiDAR and IMU in a tightly coupled manner. 
The estimated ego-motion from IMU pre-integration~\cite{IMUPRE} serves as an initial guess for scan matching LOAM-style edge and surface features and deskews the point clouds, while the LiDAR odometry estimates the bias of the IMU. 
It introduces the use of factor graphs \cite{FactorGraph} for IMU pre-integration, LiDAR odometry and any number of additional constraints, like GPS or loop closure factors. 
F-LOAM~\cite{FLOAM} combines typically separate scan-to-scan registration and scan-to-map refinement into an integrated SLAM framework where the extracted edge and surface features are matched to a local edge map and local surface map separately. 
It additionally replaces the iterative distortion correction of typical LOAM approaches with a faster two-stage method.
Similar to A-LOAM, it also makes use of the Ceres library, but applies Lie group mathematics~\cite{kirillov} to solve the ego-motion estimation problem more efficiently. 
The main limitation is that it only supports LiDAR sensors with up to 64 scan lines.
All these methods are based on point clouds but to do offer a surface representation like TSDF SLAM does.

Many robotics applications in large-scale environments benefit from a closed surface representation of the surroundings, both for navigation and localization~\cite{putz2021continuous}. 
One class of closed surface representations are Signed Distance Functions, an implicit surface representation where the signed distance is the orthogonal distance of point $x$ to the boundary of set $\Omega$ in metric space. 
The function value decreases while approaching the intersection with set $\Omega$, where the signed distance value is zero, while the sign determines whether $x$ is positioned interior or exterior of $\Omega$. 
\emph{Truncated} Signed Distance functions take this concept further and set a maximum signed distance value, clamping the magnitude of represented distance to a pre-defined to avoid inconsistencies from overlapping gradients. 
In modern SLAM approaches the kind of map representation is often neglected in the sense that the environment is only as a point cloud or voxel map. 
Even in the case of sub-sampled point clouds, the memory requirements much larger than, e.g., meshes or TSDF fields. 

In the context of TSDF and SLAM, KinectFusion~\cite{izadi_kinectfusion:_2011} uses RGB-D images for TSDF generation and presented a localization system based on so-called Projective ICP, accelerated on a GPU. 
Because KinectFusion was limited to a small volume, Kintinuous~\cite{whelan2012kintinuous} extended the approach to larger environments using a swapping strategy with a ring buffer to reduce the amount of memory allocation operations significantly.
Nießner et al.~\cite{niessner2013real} presented an efficient, alternative map representation using a spatial hashing strategy applicable for both large and fine scale volumetric reconstruction.
These related algorithms are tailored to RGB-D or time-of-flight cameras that are limited by range and suffer from lighting changes in the environment.
HATSDF-SLAM~\cite{eisoldt2022fully} overcomes range and lighting-condition limitations by adapting these ideas to LiDAR data and presented a SLAM approach that maintains a global TSDF of the environment and performs scan matching and localization directly within the signed distance field. 
It is implemented on a system-on-chip with a tightly integrated field programmable gate array (FPGA) accelerator to achieve localization in real time with a low power profile suitable for mobile systems with low energy budget.
Since this kind of hardware is difficult to program and integrate in existing systems, the aim of this work is to provide a real-time capable implementation for common embedded systems with classic CPU and embedded GPUs like NVIDIA Jetson boards.


\section{FEATURE COMPUTATION}

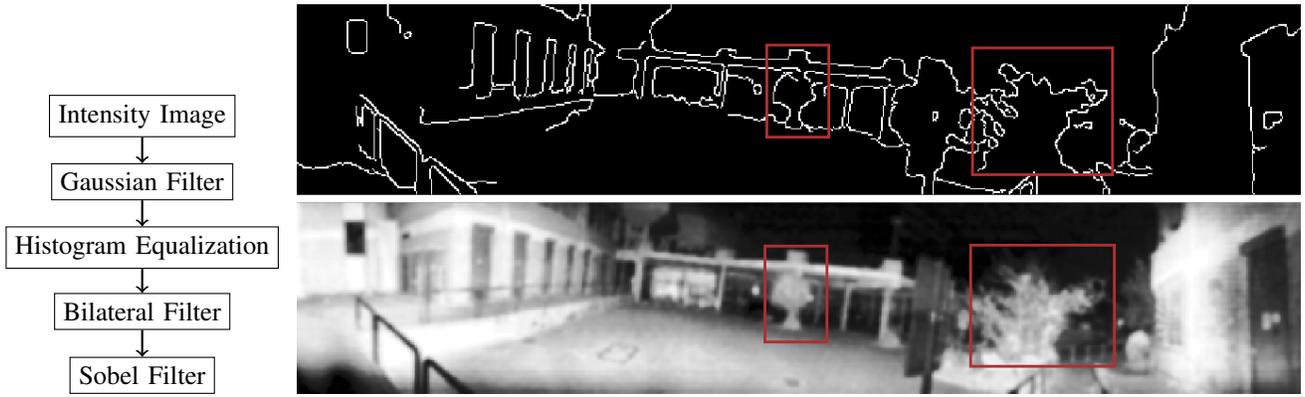
\begin{figure*}   
  \centering
  \scalebox{1.0}{\begin{tikzpicture}
	\begin{pgfonlayer}{nodelayer}
		\node [style=black rectangle] (0) at (-3, 1.125) {Gaussian Filter};
		\node [style=black rectangle] (1) at (-3, 0.25) {Histogram Equalization};
		\node [style=black rectangle] (2) at (-3, -0.6) {Bilateral Filter};
		\node [style=black rectangle] (3) at (-3, -1.45) {Sobel Filter};
		\node [style=black rectangle] (5) at (-3, 2) {Intensity Image};
	\end{pgfonlayer}
	\begin{pgfonlayer}{edgelayer}
		\draw [style=black arrow] (0) to (1);
		\draw [style=black arrow] (2) to (3);
		\draw [style=black arrow] (1) to (2);
		\draw [style=black arrow] (5) to (0);
	\end{pgfonlayer}
\end{tikzpicture}} 
  \includegraphics[width=0.75\linewidth]{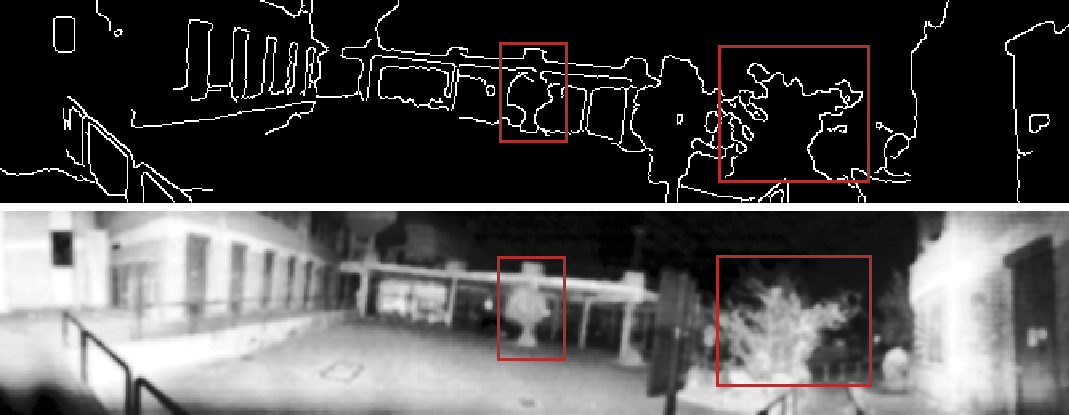} 
  \caption{Intensity image processing pipeline (left) and example image. The top right image shows the processed images after applying the filter chain on the original intensity image (bottom right). Especially around natural features like trees (marked red), much edge noise is removed.} 
  \label{fig:intensity_image_pipeline}
\end{figure*}

This section details the computation of feature points within the registration front end of \emph{FeatSense}. 
It is based on F-LOAM LiDAR mapping and uses an Ouster OS1-128 LiDAR with 128 scan lines, which is up to four times the resolution used in HATSDF SLAM. 
The official Ouster OS1-128 driver publishes a structured point cloud.
This allows to iterate over points in a very cache and memory-friendly manner to calculate the curvature $c$ of a neighborhood $S_i$, where point $P$ in line $k$ at index $i$ in the middle of this neighborhood, as shown in Eq.~\ref{eq:1}.

\begin{equation}
\label{eq:1}
  c_{(k,i)} = ((\sum_{j\in S_i, j\neq i} P_{(k, j)}) - \lvert S_i \rvert \cdot P_{(k,i)})^2 
\end{equation} 

\noindent This term is then summed across its dimensions (x,y,z) to compute the final curvature estimate. 
This follows the idea presented of F-LOAM, yet it is explicitly stated here as it strays from the original LOAM~\cite{LOAM} definition and first appeared in the implementation of A-LOAM~\cite{ALOAM}. 
After curvature calculation, feature selection follows an approach more closely in line with LOAM or LIO-SAM. 
The points in each subregion are sorted by curvature and determined to be an edge feature, if the points exceed a threshold. 
To reduce the influence of potentially incorrect correspondences further, unreliable edges found in natural objects~\cite{pointsetdiff} are discarded when exceeding a maximum edge threshold. 
Points are classified as a surface feature if they fall below a threshold and the number of selected feature points in the sub-region does not exceed a maximum.
Lastly, features are filtered out under the following conditions:
\begin{enumerate}
  \item The number of edge and surface features exceeds the maximum within a subregion.
  \item Surrounding points were already selected. As opposed to F-LOAM, this step ensures a uniform sampling of the point cloud. The OS1-128 offers a uniquely dense point cloud, so this step is crucial for odometry estimation performance. Additional sub-sampling is risky however, and relies on the assumption that changes in the environment between scans stay sufficiently small.
  \item The feature point cannot be a surface point and roughly parallel to the laser emitted from the LiDAR sensor origin. Surfaces parallel to the emitted laser pulse produce high noise, and edges can be mistaken for surfaces due to LiDAR-typical jump edge noise.
  \item The feature point cannot be on the boundary of an occluded space in the point cloud, i.e., in the shade of another object. Depending on the position of the sensor, the previously occluded region in frame $n$ can become visible in frame $n+1$ which can lead to wrong correspondences. 
\end{enumerate}

Modern LiDAR sensors include additional channels besides depth information, e.g., reflectivity, intensity or time stamps. 
Intensity is the measured return signal strength of a laser beam in a given direction.
The value can vary depending on the surface properties of the object reflecting the laser beam, scan angle, distance, texture, and humidity. 
In practice, this intensity value is often used to filter unreliable points in laser scans, as low intensity generally correlates with an imprecise measurement. 
The original F-LOAM algorithm does not take this information into account. 
In sparse regions, points are often mislabelled easily, and in more natural environments, edges may be significantly overrepresented. 
A scan of a tree, for example, has many sharp features, but depending on weather, wind or reflectivity, these types of edges are unreliable.

In FeatSense we use point density to obtain a second estimate for point cloud roughness to aid the determination of edge features.
For that, we combine the edge information obtained from an intensity image and the edge information from a LiDAR frame as computed by the LOAM algorithm.
The aim is to compute fewer but more stable edge features for motion estimation. 
Only when \textit{both} intensity image processing and LiDAR processing classify an edge or surface point they are added to the feature map.  

First, a Gaussian filter is applied to reduce noise.
As in any image, the reported intensity levels are not guaranteed to occupy the entire numerical range.
This results in loss of contrast which can hide details in darker image regions. 
Hence, we use histogram normalization to normalize the intensity image.

High-frequency noise has been shown to significantly hinder scan matching convergence~\cite{pointsetdiff}.  
To compensate for that a bilateral edge-\emph{preserving} noise reducing and smoothing filter is applied. 
In contrast to the Gaussian smoothing operator, the weights used for the calculation of pixels based on their nearby pixels depend not only on the euclidean distance but color/gray-value intensity, which ensures edge-reserving. 

Finally, the image is converted into a binary image in terms \emph{edge} or \emph{not edge} by applying a Sobel filter. 
The Sobel filter approximates the gradient of the image intensity function. 
After the image processing pipeline, high-frequency noise/features like trees are efficiently filtered, and only the silhouettes remain marked as edges, shown in Fig.~\ref{fig:intensity_image_pipeline}. 
The effect of this filter pipeline is shown in Fig.~\ref{fig:filtering}

All these steps allow parallel implementation both on GPUs and CPUs.
In FeastSense, the registration process is deliberately running on the four cores of CPU, while the embedded GPU is used for TSDF map generation.
This way, the resources of the Jetson system are exploited optimally by running the feature-based frontend with relative low bandwidth requirements on the CPU while running the more demanding TSDF update process on the GPU.

\section{FEATURE-BASED ODOMETRY ESTIMATION}

\begin{figure}
  \centering
  \scalebox{0.95}{\begin{tikzpicture}
	\begin{pgfonlayer}{nodelayer}
		\node [style=none] (2) at (-1, -0.75) {};
		\node [style=none] (3) at (-3.85, -1.75) {};
		\node [style=none] (4) at (-3, 1) {};
		\node [style=none] (5) at (-3.75, -2) {};
		\node [style=none] (6) at (-2.5, -0.5) {};
		\node [style=none] (7) at (0.5, -0.5) {};
		\node [style=none] (8) at (-0.5, -2) {};
		\node [style=small yellow dot] (9) at (-2.75, -1.5) {};
		\node [style=small yellow dot] (10) at (-2, -0.75) {};
		\node [style=small yellow dot] (11) at (-0.75, -1) {};
		\node [style=small yellow dot] (12) at (-1, -1.75) {};
		\node [style=small yellow dot] (13) at (-1.75, -1.5) {};
		\node [style=small black dot, label={above:$p_s$}] (14) at (-1, -0.5) {};
		\node [style=yellow unfilled, label={left:$p^g$}] (15) at (-1.5, -1.25) {};
		\node [style=none] (16) at (-3, -0.75) {};
		\node [style=none] (17) at (-1.5, 0.75) {};
		\node [style=none] (19) at (4.75, -0.75) {};
		\node [style=none] (20) at (2, -1.75) {};
		\node [style=none] (21) at (2.75, 1) {};
		\node [style=none] (33) at (2.75, -0.75) {};
		\node [style=small red dot] (34) at (4.075, 0.5) {};
		\node [style=small red dot] (35) at (3.975, -0.45) {};
		\node [style=small red dot] (36) at (3.9, -0.9) {};
		\node [style=small red dot] (37) at (4.075, -1.25) {};
		\node [style=small red dot] (38) at (3.9, 0.1) {};
		\node [style=small black dot, label={left:$p_e$}] (42) at (3.5, -0.25) {};
		\node [style=none] (43) at (-1, -1.25) {};
		\node [style=none] (44) at (4, -0.25) {};
		\node [style=none] (45) at (-1.5, 0.75) {};
		\node [style=none, label={left:$\vec{n_s}$}] (45) at (-1.425, 0) {};
		\node [style=none] (46) at (4, 1.4) {};
		\node [style=none] (47) at (4, -2.175) {};
		\node [style=red unfilled, label={right:$p_{e_a}$}] (48) at (4, 1) {};
		\node [style=red unfilled, label={right:$p_{e_b}$}] (49) at (4, -1.65) {};
	\end{pgfonlayer}
	\begin{pgfonlayer}{edgelayer}
		\draw [style=new edge style 0] (6.center) to (5.center);
		\draw [style=new edge style 0] (5.center) to (8.center);
		\draw [style=new edge style 0] (6.center) to (7.center);
		\draw [style=new edge style 0] (7.center) to (8.center);
		\draw [style=gray arrow] (16.center) to (4.center);
		\draw [style=gray arrow] (16.center) to (2.center);
		\draw [style=gray arrow] (16.center) to (3.center);
		\draw [style=black arrow] (15) to (17.center);
		\draw [style=black arrow] (15) to (14);
		\draw [style=gray arrow] (33.center) to (21.center);
		\draw [style=gray arrow] (33.center) to (19.center);
		\draw [style=gray arrow] (33.center) to (20.center);
		\draw [style=thick dotted green] (42) to (44.center);
		\draw [style=thick dotted green] (14) to (43.center);
		\draw [style=new edge style 0] (46.center) to (47.center);
		\draw [style=black arrow] (49) to (42);
		\draw [style=black arrow] (48) to (42);
		\draw [style=black arrow] (49) to (48);
	\end{pgfonlayer}
\end{tikzpicture}}
  \caption{Robust and modified surface patches (left) and edge correspondences (right) used in our implementation benefit from increased point cloud density. 
The distance to the respective closest surface is marked green. 
Planes are represented by the normal, $\vec{n_s}$, of the plane fitted to surface feature neighbors (yellow), while the nearest edge is represented by $p_{e_a}$ and $p_{e_b}$ projected along a fitted edge to the neighboring edge features (red). 
The respective distances are minimized in Eq.~\ref{Eq:floam_edge_mini}, \ref{Eq:floam_surface_mini} and \ref{Eq:floam_mini}} \label{fig:floam_correspondences}
\end{figure}
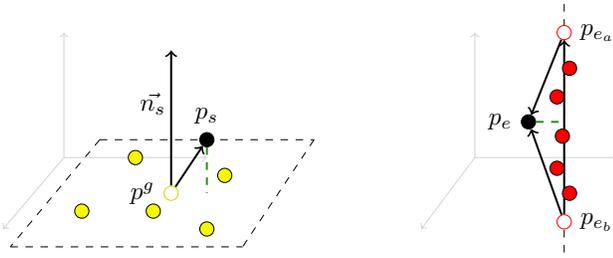

The original F-LOAM maintains an edge and surface feature map in $k$d-trees for efficient search and correspondence determination. 
In this section we explicitly describe our odometry estimation based on the original implementation\footnote{\url{https://github.com/wh200720041/floam}, accessed 08/22/2023}, which differs from the description in the original paper.

For every edge feature $p_e$, the set of 5 closest neighbors $n_e = \{n_{e_1},...,n_{e_5}\}$ are determined by a search operation in the edge feature $k$d-tree. 
The value of $n_e$ is demeaned, before calculating the covariance matrix of the set and solving for eigenvalues $\{\lambda_0, \lambda_1, \lambda_2\}$. 
The normalized eigenvector associated with the largest eigenvalue represents the unit vector $\vec{u_e}$ of the global edge formed for $n_e$. 
The global edge is considered reasonably clean if the difference between eigenvalues is large: because eigenvectors formed from eigenvalues of a covariance matrix represent the directions in which the data varies the most, differences in eigenvalues determine the spread of the data in each dimension. 
The smaller the spread, the bigger the difference in eigenvalues.  
The distance between edge feature $p_e$, given current transform $T_k$, and the global edge $e_g$ is determined by choosing $p_e$ as the geometric center of 2 nearby points $p_{e_a}$ and $p_{e_b}$ along $\vec{u_e}$.

\begin{equation}
  \label{Eq:floam_edge_mini}
  f_{e}(p_e) = \frac{\lvert (T_k p_e - p_{e_a}) \times (T_k p_{e} - p_{e_b})\rvert}{\lvert p_{e_a} - p_{e_b} \rvert}  
\end{equation}

\noindent and minimized by the Ceres optimizer. 

For every surface feature $p_s$, the set of 5 closest neighbors $n_s = \{n_{s_1},...,n_{s_5}\}$ is determined by searching the surface feature $k$d-tree. 
The normalized eigenvector associated with the smallest eigenvalue represents the normal $\vec{n_s}$ of the potential surface fitted to $n_s$. 
The quality of this fit is determined by calculating the orthogonal distance $d_{n_s}$ of neighborhood $n_s$. 
If $d_{n_s}$ falls below a threshold, the surface and surface feature point are added to the Ceres minimization problem. 
The objective minimization function for surface features $p_s$, given current transform $T_k$, surface normal $\vec{u_s}$ at position $p^g$, minimizes the surface distance

\begin{equation}
  \label{Eq:floam_surface_mini}
  f_{s}(p_s) = (T_k p_e - p^g) \cdot \vec{n_s}
\end{equation}

\noindent After finding valid correspondences for both edge and surface features, Ceres performs the optimization by minimizing the sum of weighted objective functions.

\begin{equation}
  \label{Eq:floam_mini}
  \min_{T} \sum{w(p_e)f_{e}(p_e)} + \sum{w(p_s)f_{s}(p_s)}
\end{equation}

\noindent where $w(p_e)$ and $w(p_s)$ are the edge and surface features weighted by local sharpness and smoothness, respectively. 
The optimizer is aided by an initial estimate of the pose change since the last odometry update, in this case a simple linear motion prediction. 
The odometry estimate is updated with the new transform.

\section{MAP REFINEMENT}

To further improve map quality, a post-registration step is applied. 
For that, we implemented a VICP-based optimization step, that is applied when the local map is shifted.
VGICP~\cite{VGICP} extends generalized iterative closest point (GICP) with voxelization to avoid costly nearest neighbor search. 
VGICP calculates voxel distributions from point positions instead of normals by aggregating the distribution of each point in the voxel. 
This approach allows efficient optimization in parallel. 
The authors report that the accuracy of the proposed algorithm is closely comparable to GICP, while their implementation is GPU-accelerated and outperforms GICP significantly in terms of run time. 

\begin{figure*}
  \centering
  \includegraphics[width=0.48\linewidth]{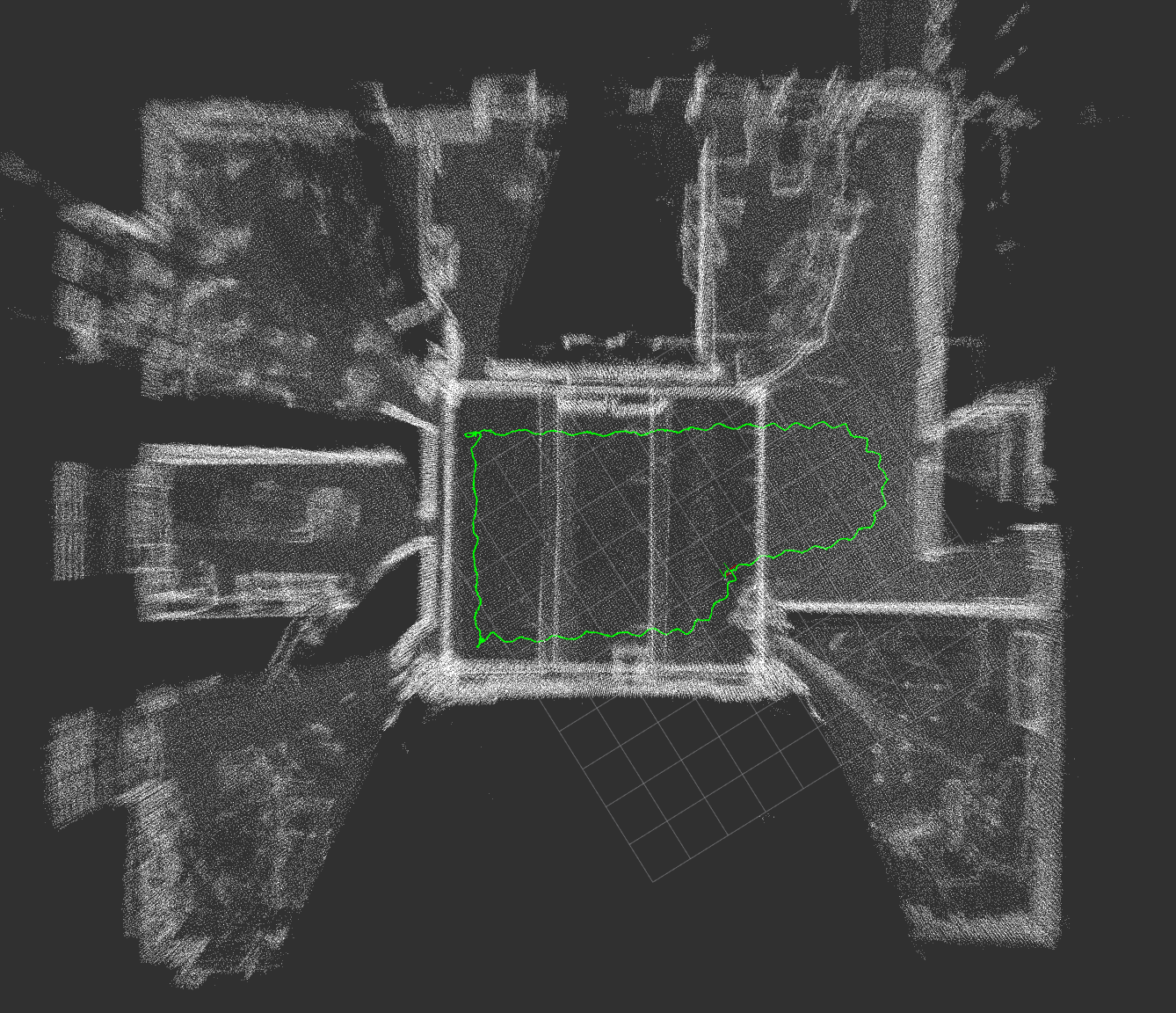}
  \hspace{1mm}
  \includegraphics[width=0.48\linewidth, trim={0 1.1cm 0 1.1cm}, clip]{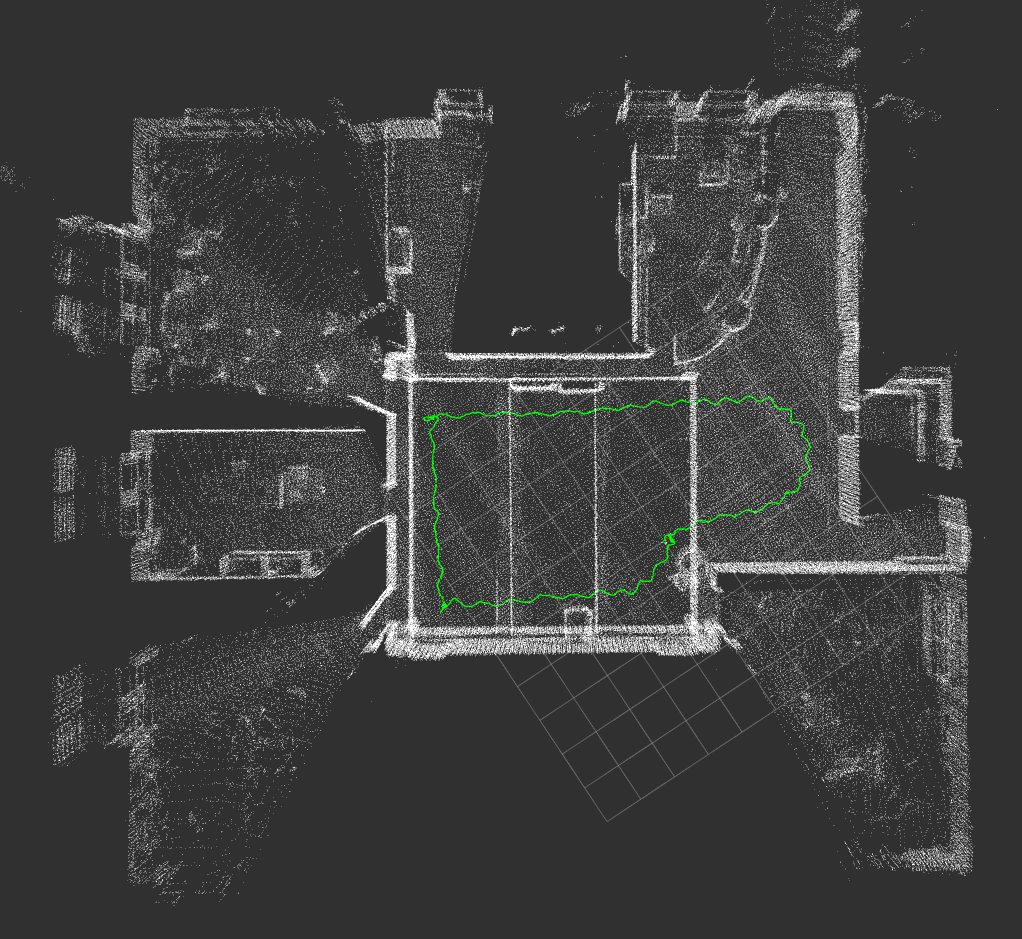}
\caption{Points of a scanned environment with feature-based matching only (left) and periodic post-processing with VGICP (right). The estimated trajectory is rendered in green. }
\label{fig:vgicp}
\end{figure*}

Although VGICP is significantly faster than GICP, it is not yet capable of running in realtime.
Hence we trigger this process only periodically.
The FeatSense odometry estimation provides an initial incremental estimate of travelled trajectory since the last optimization.
Using the incremental -- not the global state estimation -- is important, since map update has to be independent of potential drift. 
With this incremental state update, a GPU-accelerated VGICP implementation post-registers the current scan with the last $n$ registered scans in the current map, where $n$ is the number of scans taken since the last post-processing.
This post-registration results in a more consistent local map that is added to the TSDF map in the GPU mapping backend detailed in Sec.~\ref{sec:tsdf_generation}. 
After the system has moved a configurable distance (typically 5\,m), the VGICP post-registration step is triggered if real-time requirements allow so.
An impression of the effect of this step is presented in Fig.~\ref{fig:vgicp}.
Using this periodic post-processing, the drift can be reduced significantly, noticeable by the thinner walls.

\section{TSDF MAP GENERATION}
\label{sec:tsdf_generation}

This section presents the concepts and implementation of GPU-accelerated TSDF mapping in FeatSense. 
In our representation, each voxel in the local map consists of a value-weight pair.
While the value denotes the distance to the nearest surface, the weight expresses the confidence in this value. 
Because of the nature of a \emph{Truncated} Signed Distance Function, the value representing the distance to the surface is finite and can be easily fitted into 16 bits. 
The same is true for the weight, which makes it possible to reduce each voxel's memory footprint from 32 to 16 bit. 

The value-weight-pairs of the local map are filled while traversing a ray from position $p$ to scan point $x$. 
While a grid cell can intersect with multiple rays, each generated TSDF Volume $V$ chooses the smallest distance of all rays for each voxel $v$. 
The continuous mapping process, the TSDF update, then consists of averaging multiple generated TSDF volumes $V_{0}$...$V_{n}$ from positions $p_{0}$...$p_{n}$ into a consistent TSDF global volume $G$ consisting of voxels $g_{0}$...$g_{n}$. 
A single voxel $g_{i}$ is updated according to the following update rule:

\begin{equation} 
  g_{i}\text{.value} = \frac{g_{i}\text{.value} * g_{i}\text{.weight} + v_{i}\text{.value} * v_{i}\text{.weight}}{g_{i}\text{.weight} + v_{i}\text{.weight}} 
  \label{eq_1}
\end{equation}
\begin{equation} 
  g_{i}\text{.weight} = \min(\text{weight}_{\max}, g_{i}\text{.weight} + v_{i}\text{.weight}) 
  \label{eq_2}
\end{equation}

The value of $\text{weight}_{\max}$ ensures that later changes to the TSDF volume $G$ during the mapping process can have an effect on the TSDF volume. 
This becomes especially important in large datasets, where the TSDF volume can be written to from multiple scan positions in the process.

Because many modern LiDAR systems scan with a rotating LiDAR sensor with $n$ divergent horizontal scan lines, the generated TSDF volume may not be fully closed. 
Furthermore, with increasing distance of a point from the sensor origin, the gap to the closest horizontal scan line increases. 
Vertical TSDF interpolation between scan lines is implemented by calculating a sharp, elongated triangle-shaped area around the ray to scan point $p$, in which the voxels where filled with the current TSDF value during ray marching. 
To achieve this, first an interpolation vector $\vec{int}$ is calculated from a vector orthogonal to the scan lines, $\vec{up}$, and the direction vector between scanning pose $s$ and scan point $p$, $\vec{d}$.

\begin{equation}
  \vec{int} = \vec{d} \times (\vec{d} \times \vec{up})
\end{equation}

\noindent For each step $s$ in ray marching, with distance $l$ to the sensor, the height $h$ of the area that needs to be filled above and below the ray in direction $\vec{int}$ and -$\vec{int}$ is calculated. 
This is dependent of the vertical field of view $vfov$ and number of horizontal scan lines $hlines$ of the used LiDAR sensor.

\begin{equation} delta_z = tan(rad(\frac{vfov}{hlines}))\label{eq_3} \end{equation}
\begin{equation} h_s = delta_z \cdot len \label{eq_4}\end{equation} 

\noindent From here, all voxels $v_{i}$ along the interpolation vector scaled to $h_s$ are filled the current TSDF value.
The TSDF update process is illustrated in Fig.~2.

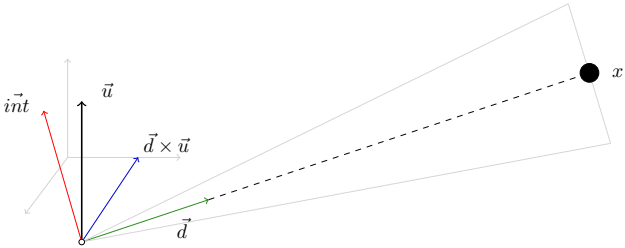
\begin{figure}
  \centering
  \scalebox{0.75}{\begin{tikzpicture}
	\begin{pgfonlayer}{nodelayer}
		\node [style=smol black dot] (0) at (0, 0) {};
		\node [style=black dot] (1) at (9, 3) {};
		\node [style=none] (2) at (1.75, 1.5) {};
		\node [style=none] (3) at (-1, 0.5) {};
		\node [style=none] (4) at (-0.25, 3.25) {};
		\node [style=none] (5) at (-0.25, 1.5) {};
		\node [style=none] (6) at (2.25, 0.75) {};
		\node [style=none] (7) at (0, 2.5) {};
		\node [style=none] (8) at (-0.675, 2.325) {};
		\node [style=none] (9) at (0.45, 2.7) {$\vec{u}$};
		\node [style=none] (10) at (-1.15, 2.475) {$\vec{int}$};
		\node [style=none] (11) at (8.625, 4.225) {};
		\node [style=none] (12) at (9.375, 1.75) {};
		\node [style=none] (13) at (1, 1.5) {};
		\node [style=none] (14) at (1.5, 1.75) {$\vec{d} \times \vec{u}$};
		\node [style=none] (15) at (9.5, 3) {$x$};
		\node [style=none] (16) at (1.775, 0.25) {$\vec{d}$};
	\end{pgfonlayer}
	\begin{pgfonlayer}{edgelayer}
		\draw [style=gray arrow] (5.center) to (4.center);
		\draw [style=gray arrow] (5.center) to (2.center);
		\draw [style=gray arrow] (5.center) to (3.center);
		\draw [style=green arrow] (0) to (6.center);
		\draw [style=new edge style 0] (6.center) to (1);
		\draw [style=black arrow] (0) to (7.center);
		\draw [style=red arrow] (0) to (8.center);
		\draw [style=thin gray line] (11.center) to (0);
		\draw [style=thin gray line] (0) to (12.center);
		\draw [style=thin gray line] (12.center) to (11.center);
		\draw [style=new edge style 2] (0) to (13.center);
	\end{pgfonlayer}
\end{tikzpicture}}
  \caption{A schematic to aid the understanding of Eq.~\ref{eq_3}. It presents the vectors that form the interpolation triangle, surrounded by gray lines, whose corresponding TSDF values are defined by Eq.~\ref{eq_1} and~\ref{eq_2}} \label{fig:interpolation_tikz}
\end{figure}

Because of the 3D discretization of space needed for a TSDF volume, multiple projective rays can calculate values for the same voxel.
The underlying data structure must therefore be accessed in a thread-safe manner. 
Because CUDA launches hundreds or thousands of threads with each kernel invocation, the CUDA run time provides a mechanism for safe, concurrent access: ``Compare and Swap'', implemented in the \texttt{atomicCAS} function. 
Compare and swap guarantees atomic access to a specific global or shared memory address. 
The \texttt{atomicCAS} function is defined as follows:
\begin{center}
\begin{small}
\texttt{uint64 atomicCAS(T *data, T oldval, T newval)}
\end{small}
\end{center}
\begin{enumerate}
  \item If “*data” is equal to “oldval”, replace it with “newval”
  \item Always returns original value of “*data”
\end{enumerate}


\begin{figure*}
  \centering
  \begin{minipage}[t]{0.28\linewidth}
  \includegraphics[width=\linewidth, trim=0 44 0 90, clip]{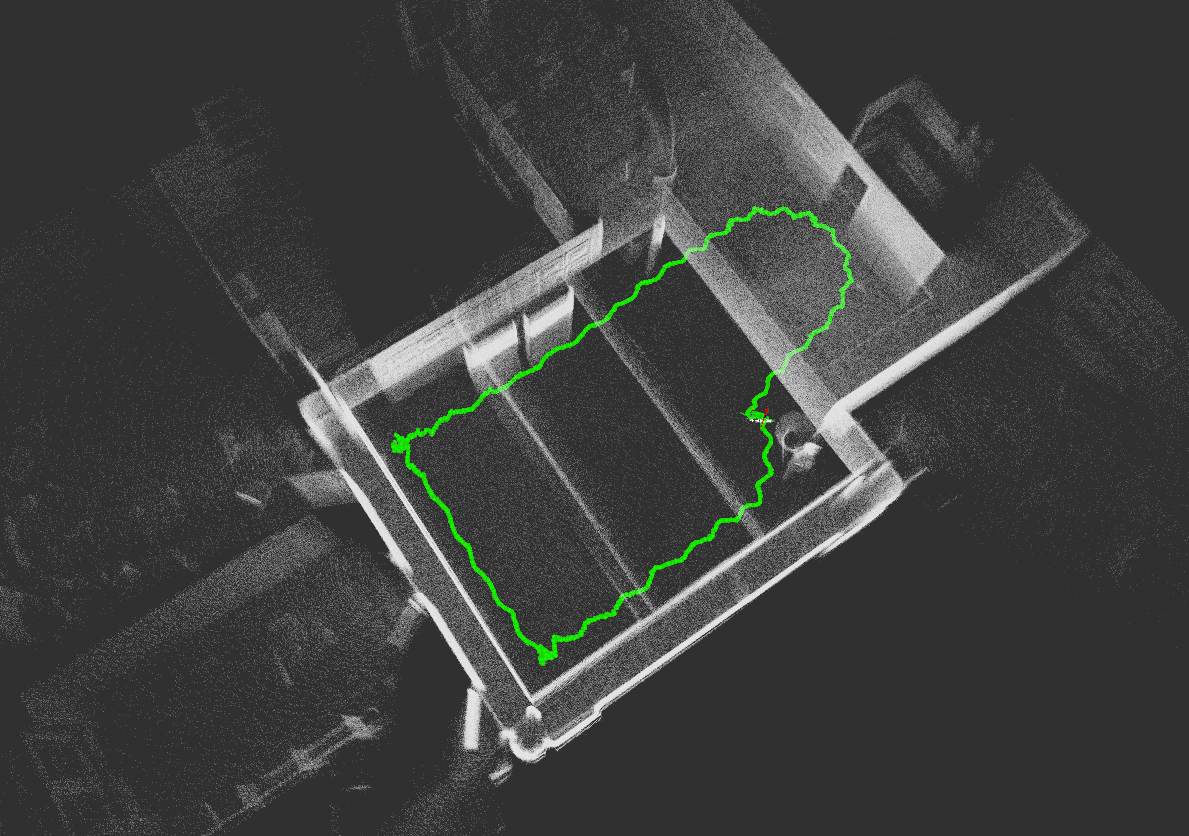}
  \includegraphics[width=\linewidth]{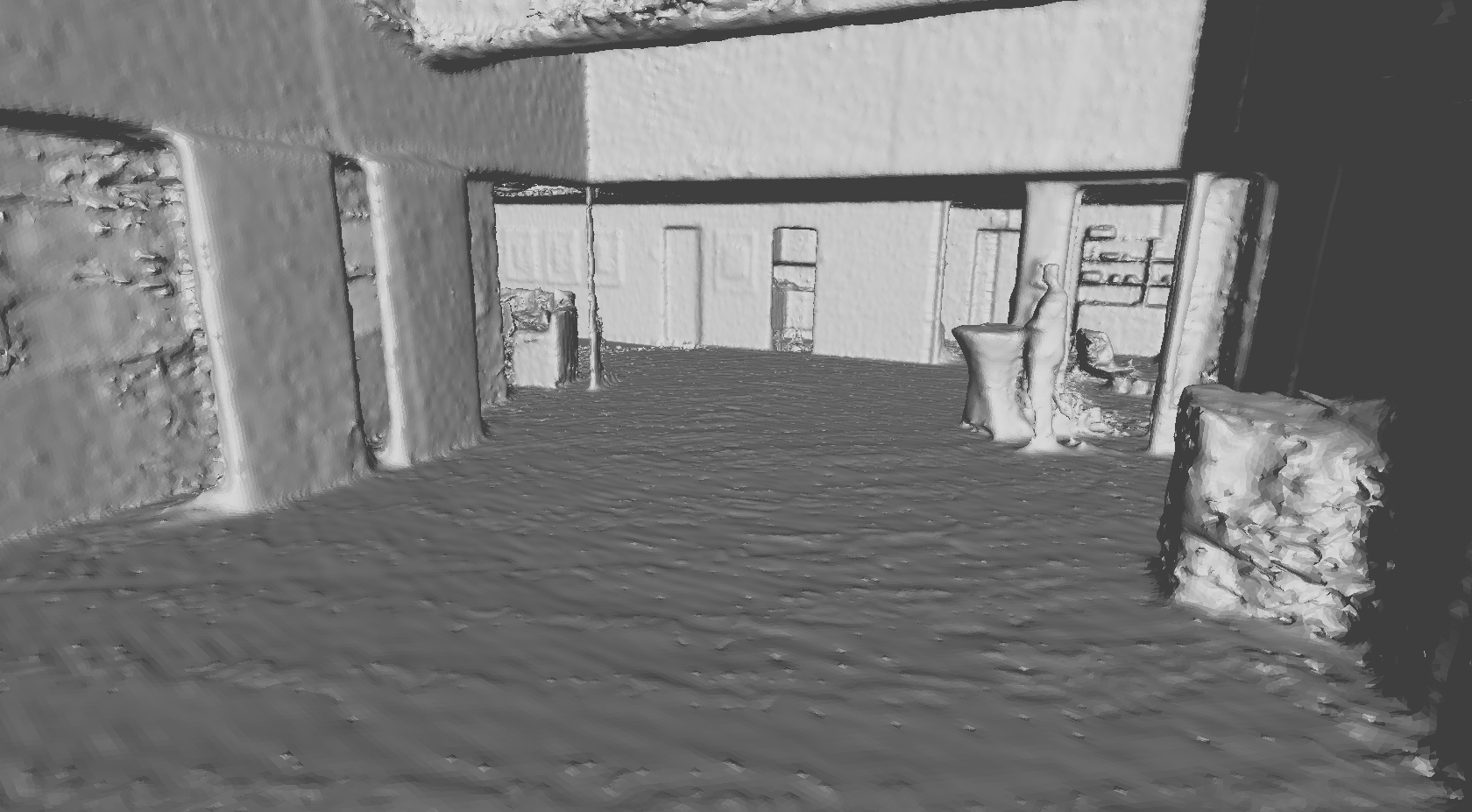}
  \end{minipage}
  \begin{minipage}{0.6\linewidth}
  \includegraphics[width=\linewidth]{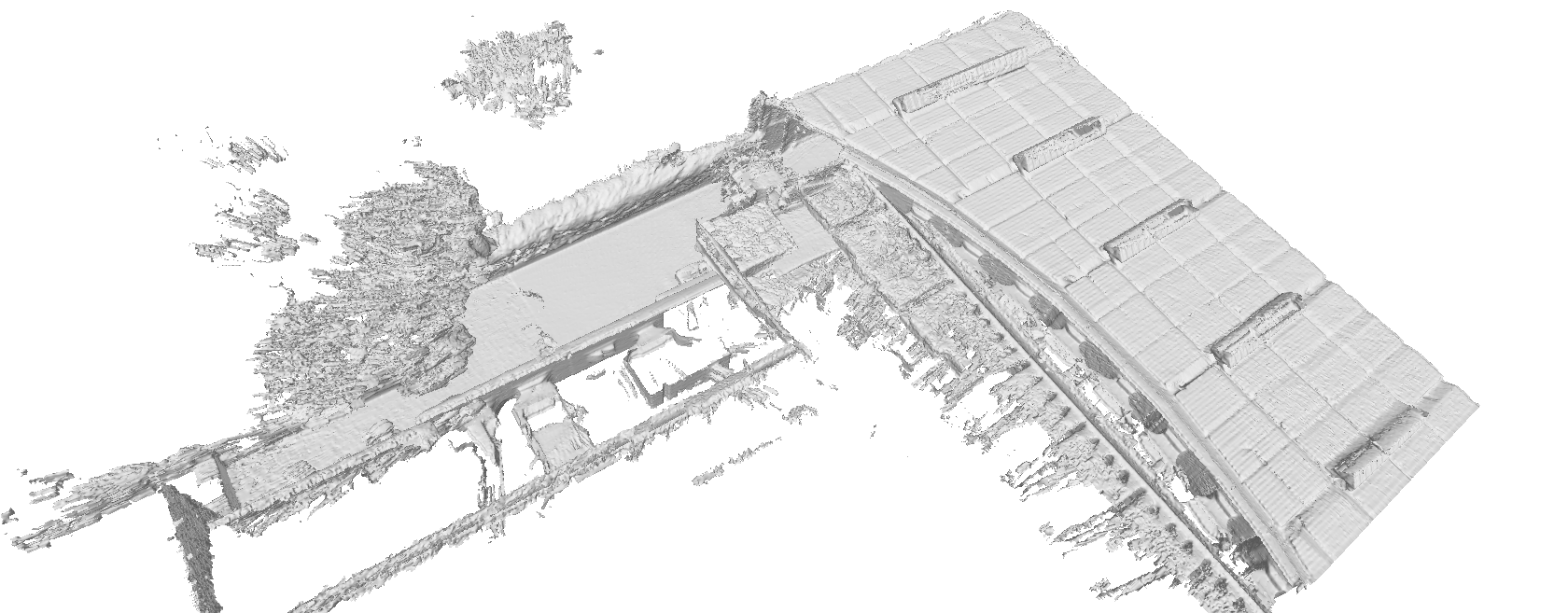}
  \end{minipage}
  \caption{Renderings of the UOS Lab (left) and the Parking Lot datasets (right).}
  \label{fig:uos_datasets}
\end{figure*}

With this concept in place, the central challenge of finding the smallest TSDF for each voxel can be solved by implementing a minimum assigning function based on \texttt{atomicCAS}. 
Each thread copies the current value in memory to a local variable, \texttt{assumed}, and \texttt{atomicCAS} returns the value in memory before swapping, \texttt{old}. 
Next, each thread compares \texttt{assumed} and \texttt{old}.

\begin{enumerate}
  \item if \texttt{assumed} and \texttt{old}, returned by \texttt{atomicCAS}, do not match, another thread has already written to memory and the thread needs to continue trying to write its potentially smallest TSDF value to memory in a loop.
  \item Each thread can break out of this loop and exit the function
  
  \begin{enumerate}
    \item if the current thread has already written to memory and therefore no swap occurs
    \item if the potential new minimum TSDF value of the thread is larger than the existing value $old$.
  \end{enumerate}
\end{enumerate}

With thread-safe access guaranteed by the \texttt{atomicCAS} directive, a continuous TSDF map can be implemented. 
To do this, two CUDA kernels have been developed: one that is responsible for generating the TSDF surface for an individual scan, \emph{min-tsdf} and another that averages the just calculated TSDF surface according to the rules presented in Eq.~\ref{eq_1}~and~\ref{eq_2}.

\section{EVALUATION}

In our evaluation, we benchmark the performance and accuracy of our work on publicly available reference datasets from the HILTI SLAM Challenge~\cite{HILTI} as well as on new datasets we previously recorded in different environments.
We tested our algorithms on different hardware (Intel PC, RTX 2080 graphics card and NVIDIA Jetson AGX Xavier board).
Because the public HILTI dataset does not feature point clouds with more than 64 scan lines, we additionally recorded two new datasets with an Ouster OS1-128, which doubles the vertical resolution~\footnote{Datasets available here: \url{http://kos.informatik.uni-osnabrueck.de/3Dscans/}}.
The new datasets are shown in Fig.~\ref{fig:uos_datasets}.
The UOS Lab dataset on the left covers an area of approximately $20\,m \times 20\,m$, the larger parking lot area is $30\,m \times 90\,m$.

\subsection{Trajectory Precision}

To evaluate the accuracy and performance of FeatSense we use the original implementation of F-LOAM as base line. 
FeatSense extracts less and more valuable features from each scan while increasing the number of Ceres optimization steps. 
Therefore, to make it a fair comparison, three configurations will be compared, each: FeatSense and F-LOAM with different number optimization steps (2 and 5).

\begin{figure*}[ht]
  \centering
  \includegraphics[width=0.32\linewidth]{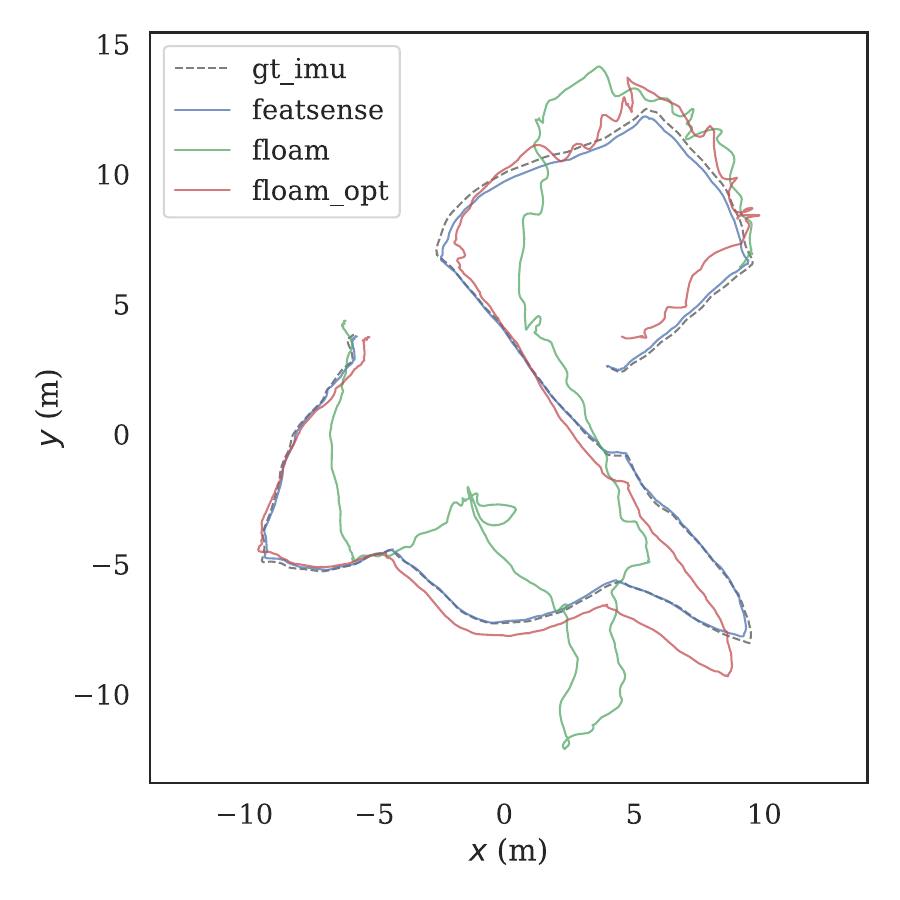}
  \includegraphics[width=0.32\linewidth]{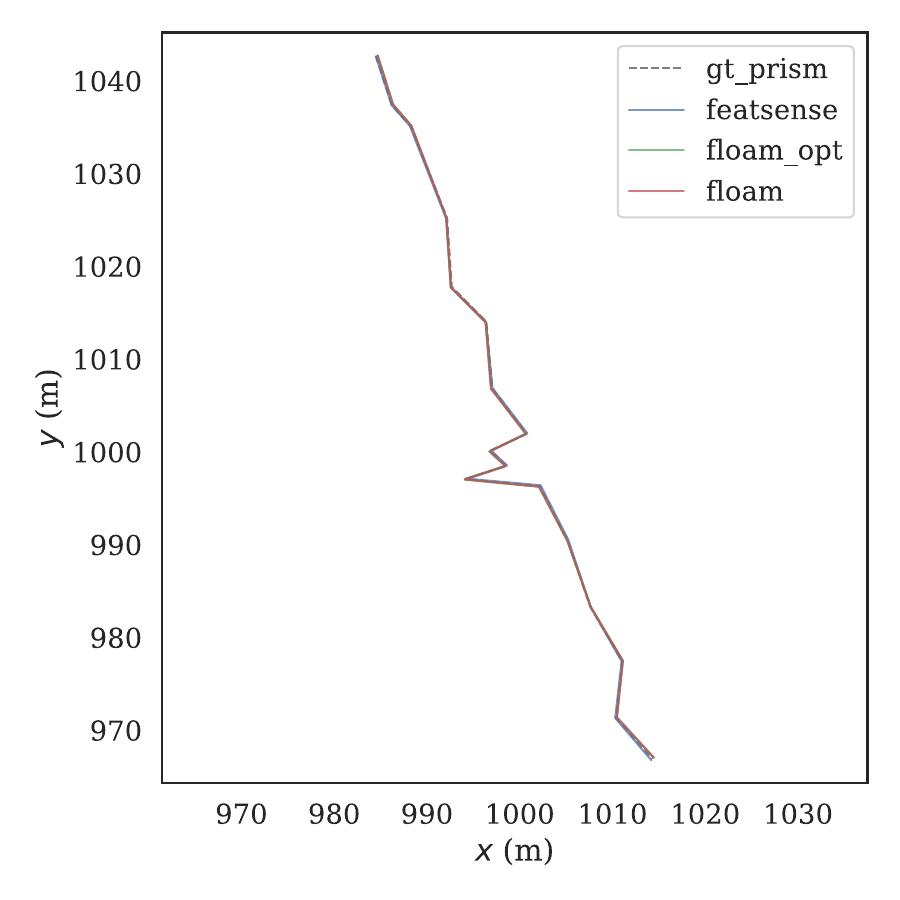}
  \includegraphics[width=0.32\linewidth]{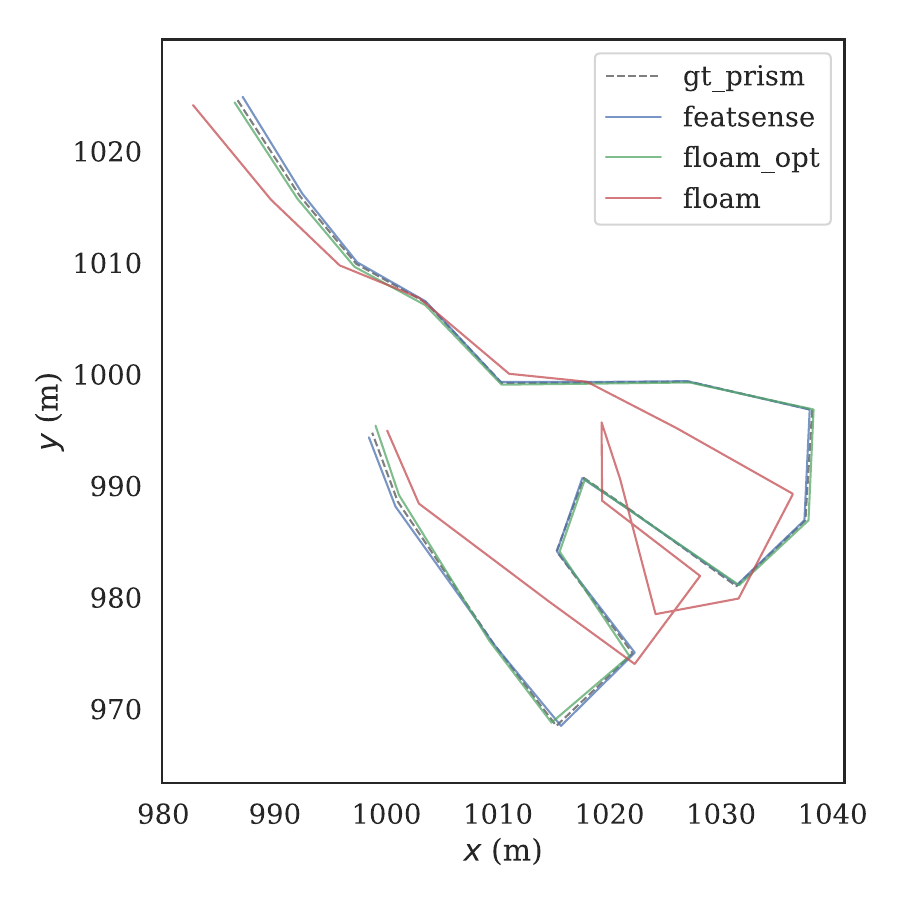}
  \caption{Comparison of the computed trajectories on the Drone (left), Basement 4 (middle) and Campus 2 (right) datasets.}
  \label{fig:hilti_results}
\end{figure*}

For precision analysis, we used data from the HILTI SLAM Challenge~\cite{HILTI}, since the setup is similar to ours. 
It is recorded in varying, challenging environments with different sensors with the intention of providing data that is considered \emph{hard}, e.g., containing movements with fast rotation changes or scans in environments with few features or high degree of similarity. 
In this evaluation, only the attached LiDAR sensor, the Ouster OS0-64, is considered. 
Data is recorded with 10 Hz, vertical resolution of 64 scan lines and 2048 lines per scan line across a vertical field of view of 90$^\circ$. 
The first dataset (Drone) was recorded in a large drone testing area with large rotation and speed changes.
The second (\emph{Basement 4}) was recorded in a small, confined basement. 
It features moderate rotation and speed changes and some ground truth markers along long corridors. 
The third (\emph{Campus 2}) is challenging, as it features unstructured environments with large open spaces and few features towards the end of the recording.
The results for established error metrics (RMSE error, mean error, standard deviation, min and max error) are summarized in Tab.~\ref{tab:hilti_results} and Fig.~\ref{fig:hilti_results}.

\begin{table}
\begin{small}
  \centering
  \caption{Trajectory errors on different HILTI benchmark datasets.}
  \label{tab:hilti_results}
  \begin{tabular}{l l l l l l l l l}
    \toprule
     Dataset  & Method & \emph{rmse} & \emph{mean} & \emph{std} & \emph{min} & \emph{max} \\
    \midrule
     Drone     & FeatSense & \textbf{0.19} & \textbf{0.17} & \textbf{0.09} & \textbf{0.01} & \textbf{0.39} \\ 
               & F-LOAM\_5  & 1.07          & 0.92          & 0.55          & 0.12          & 2.64 \\ 
               & F-LOAM\_2  & 4.25          & 3.68          & 2.12          & 0.52          & 8.22\\ 
    \midrule
    Basement 4 & FeatSense & 0.23          & 0.21          &  0.10         & 0.06          & 0.45 \\ 
               & F-LOAM\_5  & 0.15          & 0.14          & \textbf{0.06} & 0.05          & \textbf{0.28} \\ 
               & F-LOAM\_2  & \textbf{0.15} & \textbf{0.13} & 0.07          & \textbf{0.04} & 0.31 \\ 
    \midrule
    Campus 2   & FeatSense & \textbf{0.28} & \textbf{0.24} & \textbf{0.15} & 0.05          & \textbf{0.58} \\ 
               & F-LOAM\_5  & 0.35          & 0.30          & 0.19          & \textbf{0.04} & 0.73  \\ 
               & F-LOAM\_2  & 5.73          & 4.80          & 3.13          & 0.47          & 9.49  \\ 
    \bottomrule
    \end{tabular}
  \end{small}
\end{table}

\begin{table}
  \begin{small}
   \centering
   \caption{Registration run times in ms at full resolution on a PC and the ARM CPU of the Jetson board.}
   \label{tab:slam_performance}
   \begin{tabular}{llcc}
  \toprule
                    & & UOS Lab &  UOS Parking Lot \\
                    & & Intel / ARM  & Intel / ARM\\
   \midrule
  Pre-proc.     &  F-LOAM & 8 / 26 & 18 / 61 \\
                    &  FeatSense & \textbf{6} / \textbf{10} & \textbf{8} / \textbf{17} \\
  \midrule
  Registration      & F-LOAM & 15 / 32 & 79 / 197 \\
                    & FeatSense      & \textbf{7}/ \textbf{16} & \textbf{34} / \textbf{76} \\
   \midrule
   Total            & F-LOAM & 23 / 58 & 97 / 257 \\
                    & FeatSense & \textbf{13} / \textbf{26} & \textbf{42} / \textbf{93} \\
              \bottomrule
   
  \end{tabular}
\end{small}
\end{table}

Except for the Basement~4 dataset, FeatSense outperforms F-LOAM in terms of accuracy with respect to the ground truth trajectory provided by the HILTI datasets.
It has to be noted that in datasets tracked with prism total station tracker, a large drift using benchmarking tools is observed at the very end and start of the trajectory plots.
However, the map is consistent at both start and end position in all datasets with prism ground truth, and neither F-LOAM nor FeatSense have problems with registration at these positions. 
This may explain the measured weakness in the Basement~4 dataset, but investigation is needed to determine weather this is a problem with the benchmarking or a problem case for FeatSense.
The overall evaluation on all datasets indicates that FeatSense delivers accurate trajectories. 

\subsection{Registration and Preprocessing Benchmarks}

FeatSense's run time was evaluated using an Ouster OS1-128 running at full resolution of 1024x128 at recording speeds between 10 and 20\,Hz. 
We measured the time for pre-processing and registration individually to demonstrate the benefits from the ordered structure of the incoming data compared to F-LOAM, where re-ordering is required.
On the smaller dataset, FeatSense is twice as fast as F-LOAM, on the larger dataset the speedup is even higher.
The results also show that our implementation is capable to integrate data at 10\,Hz (the lower sensor frequency) on the Jetson board, on the Intel computer real time integration in the 20\,Hz mode is possible.

\subsection{TSDF Performance}

To meaningfully compare against a reference implementation, we created an OpenMP-multithreaded CPU-implementation of HATSDF-SLAM~\cite{eisoldt2022fully}. 
This reference was compared to the GPU TSDF implementation on the NVIDIA Jetson AGX Xavier in different power modes and a PC with an NVIDIA RTX 2080 graphics card.
In lower power modes, the number of available GPU resources and CPU clocking is limited to reduce the power consumption.
To investigate, if the current implementation can be applied on Jetson boards with less resources, we benchmarked the available settings provided by the Jetson SDK. 
Additionally, we compared our TSDF algorithm against the original FPGA-accelerated implementation on the Xilinx Zynq UltraScale+ ZU15EG presented in~\cite{eisoldt2022fully}. 
The voxel size of the TSDF grid was set to 6.4\,cm, which provides a good balance between performance and map accuracy, as evaluated in previous work~\cite{eisoldt2022fully}, where we compared the meshes generated from the TSDF representation with manually measured ground truth.
The OS1-128 sensor was again operated at full resolution.
The results are summarized in Tab.~\ref{tab:gpu_runtime_tsdf}.
Note that the larger local map used in the second experiments exceeded the available resources on the FPGA-board.

\begin{table}
\centering
\begin{small}
\caption{TSDF run times in ms for both UOS lab dataset and UOS parking lot datasets with different local map sizes with a voxel resolution of 6.4\,cm.}
\label{tab:gpu_runtime_tsdf}
\begin{tabular}{lcccccc}
  \toprule
  & \multicolumn{3}{c}{\textbf{20x20x15m, 6.4cm}} & \multicolumn{3}{c}{\textbf{40x40x15m, 6.4cm}}
  \\\cmidrule(lr){2-4}\cmidrule(lr){5-7}
  & \emph{min} & \emph{max} & \emph{avg}    & \emph{min} & \emph{max} & \emph{avg}\\\midrule
    i7-4790K x 8 & 55 & 1018 & 407 & 1932 & 3214 & 2752 \\
    Xilinx ZU15EG & 11 & 557 & 104 & n.a. & n.a. & n.a. \\
    RTX 2080 & \textless 1 & \textless 1 & \textless 1 & \textless 1 & \textless 1 & \textless 1 \\
    Xavier M0/MAX & \textless 1 & \textless 1 & \textless 1 & \textless 1 & 16 & \textless 1\\
    Xavier M1/10W & \textless 1 & 14 & 1 & \textless 1 & 36 & 2\\
    Xavier M2/15W & \textless 1 & 2 & \textless 1 &  \textless 1 & 47 & 3\\
    Xavier M3/30W & \textless 1  & \textless 1 & \textless 1 & \textless 1 & 24 & 1\\
    Xavier M4/30W &\textless 1 & 1 & \textless 1 & \textless 1 & 35 & 1\\
    Xavier M5/30W & \textless 1 & \textless 1 & \textless 1 & \textless 1 & 16 & \textless 1\\
    Xavier M6/30W & \textless 1 & 13 & \textless 1 & \textless 1 & 41 & 1\\
    Xavier M7/15W & \textless 1 & 1 & \textless 1 & \textless 1 & 1 & \textless 1\\
    \bottomrule
\end{tabular}
\vspace{-4mm}
\end{small}
\end{table}

The results show that the GPU implementation outperforms the original implementation typically by two orders of magnitude.
It also allows using larger local maps than the FPGA-baseline, which is desirable to reduce the I/O load while storing the global map to the SSD hard drive.
The accuracy of the meshes extracted from the UOS lab datasets were additionally compared to the meshes evaluated in~\cite{eisoldt2022fully}.
The reported mean error in CloudCompare was close to zero, which indicates that the resulting TSDF maps reflect the accuracy of this previous work.

\section{CONCLUSION AND FUTURE WORK}

In this paper, we presented FeatSense, an approach to build LiDAR-based TSDF maps in embedded devices.
In contrast to existing methods, FeatSense uses information from the intensity images of modern LiDARs to generate more stable features for scan matching.
A simple yet effective image processing pipeline filters out instable features and reduces the number of features needed to solve the registration problem, which in turn decreases computation time.
FeatSense outperforms F-LOAM in terms of processing speed and accuracy.
The presented results show, that this approach allows processing of high-resolution point clouds from an Ouster OS1-128 in real time on a Jetson AGX Xavier board.
The TSDF mapping backend runs on a GPU and is able to integrate more data faster than the previous implementation in HATSDF SLAM and avoids the I/O bottleneck by supporting larger local maps.
In future work, we plan to port the TSDF-based registration used in HATSDF SLAM to GPUs to also the computed TSDF map directly for scan registration.

\section*{ACKNOWLEDGMENT}

The DFKI Niedersachsen (DFKI NI) is sponsored by the Ministry of Science and Culture of Lower Saxony and the VolkswagenStiftung

\bibliographystyle{IEEEtran}
\bibliography{IEEEabrv, papers}

\begin{thebibliography}{10}
\providecommand{\url}[1]{#1}
\csname url@samestyle\endcsname
\providecommand{\newblock}{\relax}
\providecommand{\bibinfo}[2]{#2}
\providecommand{\BIBentrySTDinterwordspacing}{\spaceskip=0pt\relax}
\providecommand{\BIBentryALTinterwordstretchfactor}{4}
\providecommand{\BIBentryALTinterwordspacing}{\spaceskip=\fontdimen2\font plus
\BIBentryALTinterwordstretchfactor\fontdimen3\font minus
  \fontdimen4\font\relax}
\providecommand{\BIBforeignlanguage}[2]{{%
\expandafter\ifx\csname l@#1\endcsname\relax
\typeout{** WARNING: IEEEtran.bst: No hyphenation pattern has been}%
\typeout{** loaded for the language `#1'. Using the pattern for}%
\typeout{** the default language instead.}%
\else
\language=\csname l@#1\endcsname
\fi
#2}}
\providecommand{\BIBdecl}{\relax}
\BIBdecl

\bibitem{eisoldt2022fully}
M.~Eisoldt, J.~Gaal, T.~Wiemann, M.~Flottmann, M.~Rothmann, M.~Tassemeier, and
  M.~Porrmann, ``A fully integrated system for hardware-accelerated tsdf slam
  with lidar sensors (hatsdf slam),'' \emph{Robotics and Autonomous Systems},
  vol. 156, p. 104205, 2022.

\bibitem{LOAM}
J.~Zhang and S.~Singh, ``Loam : Lidar odometry and mapping in real-time,''
  \emph{Robotics: Science and Systems Conference (RSS)}, pp. 109--111, 01 2014.

\bibitem{ALOAM}
\BIBentryALTinterwordspacing
H.~A.~R. Lab. Advanced lidar odometry and mapping (a-loam). Accessed last
  11/15/22. [Online]. Available:
  \url{https://github.com/HKUST-Aerial-Robotics/A-LOAM}
\BIBentrySTDinterwordspacing

\bibitem{CERES}
\BIBentryALTinterwordspacing
S.~Agarwal, K.~Mierle, and T.~C.~S. Team, ``{Ceres Solver},'' 3 2022. [Online].
  Available: \url{https://github.com/ceres-solver/ceres-solver}
\BIBentrySTDinterwordspacing

\bibitem{LEGOLOAM}
T.~Shan and B.~Englot, ``Lego-loam: Lightweight and ground-optimized lidar
  odometry and mapping on variable terrain,'' in \emph{2018 IEEE/RSJ
  International Conference on Intelligent Robots and Systems (IROS)}.\hskip 1em
  plus 0.5em minus 0.4em\relax IEEE, 2018, pp. 4758--4765.

\bibitem{LIOSAM}
T.~Shan, B.~Englot, D.~Meyers, W.~Wang, C.~Ratti, and D.~Rus, ``Lio-sam:
  Tightly-coupled lidar inertial odometry via smoothing and mapping,'' in
  \emph{2020 IEEE/RSJ International Conference on Intelligent Robots and
  Systems (IROS)}, 2020, pp. 5135--5142.

\bibitem{IMUPRE}
C.~Forster, L.~Carlone, F.~Dellaert, and D.~Scaramuzza, ``Imu preintegration on
  manifold for efficient visual-inertial maximum-a-posteriori estimation,'' in
  \emph{Robotics: Science and Systems}, 2015.

\bibitem{FactorGraph}
D.~Koller and N.~Friedman, \emph{Probabilistic graphical models: principles and
  techniques}.\hskip 1em plus 0.5em minus 0.4em\relax MIT press, 2009.

\bibitem{FLOAM}
H.~{Wang}, C.~{Wang}, C.~{Chen}, and L.~{Xie}, ``F-loam : Fast lidar odometry
  and mapping,'' in \emph{2021 IEEE/RSJ International Conference on Intelligent
  Robots and Systems (IROS)}, 2020.

\bibitem{kirillov}
A.~Kirillov, Jr, \emph{An Introduction to Lie Groups and Lie Algebras}, ser.
  Cambridge Studies in Advanced Mathematics.\hskip 1em plus 0.5em minus
  0.4em\relax Cambridge University Press, 2008.

\bibitem{putz2021continuous}
S.~P{\"u}tz, T.~Wiemann, M.~K. Piening, and J.~Hertzberg, ``Continuous shortest
  path vector field navigation on 3d triangular meshes for mobile robots,'' in
  \emph{2021 IEEE International Conference on Robotics and Automation
  (ICRA)}.\hskip 1em plus 0.5em minus 0.4em\relax IEEE, 2021, pp. 2256--2263.

\bibitem{izadi_kinectfusion:_2011}
S.~Izadi, R.~A. Newcombe, D.~Kim, O.~Hilliges, D.~Molyneaux, S.~Hodges,
  P.~Kohli, J.~Shotton, A.~J. Davison, and A.~Fitzgibbon, ``{KinectFusion}:
  {Real}-time {Dynamic} 3d {Surface} {Reconstruction} and {Interaction},'' in
  \emph{{ACM} {SIGGRAPH} 2011 {Talks}}.\hskip 1em plus 0.5em minus 0.4em\relax
  ACM, 2011.

\bibitem{whelan2012kintinuous}
T.~Whelan, M.~Kaess, M.~Fallon, H.~Johannsson, J.~Leonard, and J.~McDonald,
  ``Kintinuous: Spatially extended {K}inect{F}usion,'' in \emph{RSS Workshop on
  RGB-D: Advanced Reasoning with Depth Cameras}, Sydney, Australia, Jul 2012.

\bibitem{niessner2013real}
M.~Nie{\ss}ner, M.~Zollh{\"o}fer, S.~Izadi, and M.~Stamminger, ``Real-time 3d
  reconstruction at scale using voxel hashing,'' \emph{ACM Transactions on
  Graphics (ToG)}, vol.~32, no.~6, pp. 1--11, 2013.

\bibitem{pointsetdiff}
\BIBentryALTinterwordspacing
Q.~R. Graehling. Feature extraction based iterative closest point registration
  for large scale aerial lidar point cloud. Accessed last 11/16/22. [Online].
  Available: \url{https://docs.nvidia.com/cuda/profiler-users-guide/index.html}
\BIBentrySTDinterwordspacing

\bibitem{VGICP}
K.~Koide, M.~Yokozuka, S.~Oishi, and A.~Banno, ``Voxelized gicp for fast and
  accurate 3d point cloud registration,'' EasyChair Preprint no. 2703,
  EasyChair, 2020.

\bibitem{HILTI}
M.~Helmberger, K.~Morin, B.~Berner, N.~Kumar, D.~Wang, Y.~Yue, G.~Cioffi, and
  D.~Scaramuzza, ``The hilti slam challenge dataset,'' 2021.

\end{thebibliography}

\balance

\end{document}